**Context Matters: Comparison of commercial large language tools in veterinary medicine.**

Tyler J Poore[1], Christopher J Pinard[1,2,3,4], Aleena Shabbir[1], Andrew Lagree[1], Andre Telfer[1,5], Kuan-Chuen Wu[1,6]

1. ANI.ML Research, ANI.ML Health Inc., Toronto, Ontario, Canada
2. Department of Clinical Studies, University of Guelph, Guelph, Ontario, Canada
3. Centre for Responsible and Ethical Artificial Intelligence (CARE-AI), University of Guelph, Guelph, Ontario, Canada
4. Toronto Animal Cancer Centre, Toronto, Ontario, Canada
5. The Hospital for Sick Children, Toronto, Ontario, Canada
6. Department of Mechatronic Engineering, Anderson University, Anderson, Indiana, USA

**Author Approval:** All authors have seen and approved the manuscript.

**Competing Interests:** All authors acknowledge their direct affiliation with ANI.ML Health and the development of the veterinary language tool, Hachiko (listed as Product 1 in this manuscript).

**Declarations:** There are no further competing interests to declare.

**Data Availability Statement:** The de-identified data for this study is available upon request for non-commercial purposes. The methodology is available through the ANI.ML Research GitHub repository and is publicly available at:
https://github.com/pooret-animl/vet-summary-eval-framework

**Funding Statement:** No funding was provided for the execution of this project.

**Abstract:**
Large language models (LLMs) are increasingly used in clinical settings, yet their performance in veterinary medicine remains underexplored. We evaluated three commercially available veterinary-focused LLM summarization tools—Product 1 (Hachiko) and Products 2 and 3—on a standardized dataset of veterinary oncology records. Using a rubric-guided LLM-as-a-judge framework, summaries were scored across five domains: Factual Accuracy, Completeness, Chronological Order, Clinical Relevance, and Organization. Product 1 achieved the highest overall performance, with a median average score of 4.61 (IQR: 0.73), compared to 2.55 (IQR: 0.78) for Product 2 and 2.45 (IQR: 0.92) for Product 3. It also received perfect median scores in Factual Accuracy and Chronological Order. To assess the internal consistency of the grading framework itself, we repeated the evaluation across three independent runs. The LLM grader demonstrated high reproducibility, with Average Score standard deviations of 0.015 (Product 1), 0.088 (Product 2), and 0.034 (Product 3). These findings highlight the importance of veterinary-specific commercial LLM tools and demonstrate that LLM-as-a-judge evaluation is a scalable and reproducible method for assessing clinical NLP summarization in veterinary medicine.

**Introduction:**
The rapid advancement of large language models (LLMs) has sparked considerable interest across various professional fields, including healthcare.[1–8] These tools offer potential benefits such as improved efficiency in information retrieval, decision support, and even automated content generation.[9–15] Veterinary medicine stands to benefit from the application of LLMs. However, the efficacy of these models is heavily reliant on their ability to understand and generate language accurately within the specific nuances of veterinary terminology, clinical scenarios, and ethical considerations.[16,17] We previously attempted to directly adapt a human medical NER model to veterinary clinical records, showing very poor adaptability of the model to a veterinary context and underscoring the need for veterinary-specific models.[18] While several commercial platforms that utilize LLMs are emerging for veterinary applications (scribes, summarization tools, clinical decision support and chatbots), a comprehensive framework for comparative analysis of their performance remains unavailable to the veterinary team and is largely unexplored in the profession.

At present, there are no veterinary-trained language models. Consequently, platforms depend on either prompt engineering of publicly available models, integration with validated vector databases (e.g., PetBERT), or a hybrid of these methods.[19] In human medicine, LLMs have served as reliable evaluators (judges) of model adaptability in clinical contexts.[2] In this scenario, an agentic framework allows LLMs to function as agents, assessing clinical text outputs according to parameters specified by human observers.

This study aimed to evaluate and compare three commercially available LLM-powered tools across key criteria relevant to clinical utility in veterinary medicine. We hypothesized that differences in response quality would arise based on each platform's pipeline, particularly given that many of these tools, to the authors' knowledge, have not been developed with veterinary-specific training. Here, we describe our methodology for assessing these tools using an LLM-as-a-judge framework and provide a comparative analysis of their performance, contributing to a more nuanced understanding of LLM capabilities and limitations within the veterinary domain.

**Methods:**
Medical records were obtained from a private data repository hosted on a HIPAA compliant Amazon Web Services (AWS) server located in Canada. The medical records were all from a single domain (medical oncology). Each clinical record was submitted to each commercial summarization platform, capable of ingestion and summarization of clinical records. Each platform's "Standard" summary option was selected for consistency. Where custom instructions were allowed to be integrated into the clinical record, we used a standard instruction: "Could you provide a detailed medical history of the patient, including the age, breed, all diagnoses, bloodwork, and test results?." All code for the grading framework is publicly available (https://github.com/pooret-animl/vet-summary-eval-framework) to support reproducibility and further research.

Each of the 42 clinical records was processed through three commercial summarization platforms capable of ingesting and summarizing veterinary clinical records: (1) Product 1 (Hachiko): A proprietary veterinary-specific language model pipeline developed in-house with domain-specific training on veterinary medical data, and two other veterinary commercial platforms (2 & 3) currently available with the reported capabilities of PDF medical record summarization.

We developed an automated grading system using Google's Gemini 2.5 Pro model as an impartial judge to evaluate summary quality. The framework was designed to minimize bias and ensure reproducible evaluation across all summaries. The system employed Gemini 2.5 Pro with structured output capabilities, configured with a temperature of 0.1, JSON response formatting with Pydantic schema validation, and a reasoning budget of 16,384 tokens. The large reasoning budget was chosen to give the model sufficient space to reason through difficult scenarios and to reduce the likelihood of snap judgments.

The LLM judge evaluated each summary against five weighted criteria that were developed in consultation with a board-certified veterinary clinician. Each criterion was scored on a scale from 1 (Poor) to 5 (Excellent). Factual accuracy (weight: 2.5) measured the alignment of specific facts such as dates, patient identifiers, diagnoses, treatments, and test results between the summary and the source text, with mismatches or fabrications explicitly identified. Completeness (weight: 1.2) assessed whether key medical events, diagnoses, treatments, and significant findings were included in the summary, noting major omissions. Chronological order (weight: 1.0) evaluated whether the temporal sequence of events in the summary reflected the original timeline. Clinical relevance (weight: 1.5) examined whether the summary emphasized medically important information appropriate for referral or medical history, while avoiding trivial details that detracted from context. Organization (weight: 0.8) focused on structure, clarity, and logical flow, including whether information was arranged chronologically or by problem list.

The grading process followed a structured workflow. First, the LLM judge performed detailed internal reasoning for each criterion before assigning a score, ensuring transparency in its decision-making. Next, scores ranging from 1 to 5 were assigned for each criterion along with written feedback. A final weighted score was then calculated using the formula:

$$\text{Weighted Score} = \frac{\sum(score_i \times weight_i)}{\sum(weight_i)}$$

where *i* denotes each evaluation criterion. To ensure reliability, JSON schema validation was enforced for all LLM outputs, and failed grading attempts triggered automatic retries. Each grading session included the full source text to prevent reliance on external knowledge; no text truncation or retrieval-augmented generation was required.

Results were aggregated into structured reports containing criterion-level scores and feedback for each summary, weighted average scores per platform, distribution analyses using box plots to visualize variability, and heatmap visualizations of median scores by category and platform. Median values and interquartile ranges (IQR) were reported for each platform to capture both central tendency and consistency. The median was chosen over the mean as a non-parametric measure more appropriate for ordinal scales and for mitigating the influence of outliers.

To ensure validity in the LLM-as-a-judge framework we also ran the entire dataset in triplicate and evaluated the standard deviations of the scoring outputs.

**Results:**
Product 1 achieved the highest overall performance, with a median weighted score of 4.61, compared to 2.55 for Commercial Platform 2 and 2.45 for Commercial Platform 3 (**Figure 1**). The interquartile range (IQR) for Product 1 was notably smaller, suggesting greater consistency in performance across outputs. In contrast, Commercial Platform 3 displayed a wider spread of scores, including one high-scoring outlier, which indicates that while it is capable of producing strong responses, these were inconsistent and accompanied by a higher risk of low-quality outputs.

When median scores and IQRs were analyzed in detail, Commercial Platform 3 achieved a median of 2.45 with an IQR of 0.92, Commercial Platform 2 achieved a median of 2.55 with an IQR of 0.78, and Product 1 achieved a median of 4.61 with an IQR of 0.73. These results reinforce that Product 1 not only outperformed the other platforms but did so with greater reliability.

**Figure 2** presents a heatmap illustrating the average scores by category and platform. Product 1 consistently achieved the highest average values across all categories: Chronological Order (5.0), Clinical Relevance (4.0), Completeness (4.0), Factual Accuracy (5.0), and Organization (5.0). In contrast, Commercial Platform 2 demonstrated only moderate performance. Commercial Platform 3 consistently underperformed relative to both Product 1 and Commercial Platform 2, particularly in Chronological Order and Organization, suggesting significant challenges in maintaining logical structure. Across all platforms, factual accuracy was a persistent challenge, but Product 1 consistently outperformed the others in this critical dimension.

To evaluate the internal reproducibility of the LLM-based grading framework, we assessed the standard deviation of scores assigned by the grader across three independent evaluations of the same summaries. As shown in **Figure 4**, the grading tool demonstrated high internal consistency, with low standard deviation observed across all platforms and categories. The standard deviation of Average Score was just 0.015 for summaries generated by Product 1, compared to 0.088 for Product 2 and 0.034 for Product 3. Similarly, across individual categories such as Factual Accuracy (0.036 for Product 1 vs. 0.113 and 0.143 for Products 2 and 3, respectively) and Chronological Order (0.036 for Product 1 vs. 0.172 and 0.084), the grader's evaluations remained remarkably stable for a given input. These results suggest that the LLM grader produces reproducible, stable assessments with minimal intra-tool variability, bolstering confidence in its utility for benchmarking and evaluation of veterinary LLM-generated summaries.

**Discussion:**

This comparative analysis highlights substantial differences in the performance of language model platforms when applied to veterinary medical contexts, specifically for summarization tasks. The similarities observed between the two commercial platforms raise concerns for the use of similar models across platforms that may share limitations, including insufficient domain-specific training and a lack of robust guardrails.

By contrast, the consistently high scores achieved by Product 1 demonstrate the benefits of a pipeline explicitly trained and designed for/on veterinary data. This finding underscores the importance of developing transparent, domain-specific models/pipelines rather than relying on generalized LLM outputs. At minimum, the use of carefully engineered prompting strategies and targeted post-processing is necessary to ensure clinically useful content when employing wrapper-based approaches around general-purpose LLMs. The smaller IQR observed for Product 1 also reflects greater reliability and predictability, which is essential for clinical deployment.

Commercial Platform 2 exhibited variable performance, with lower scores in Chronological Order and Organization, suggesting that it may require improved prompting or post-processing to ensure clarity and logical flow. The presence of high-performing outliers indicates that the model is capable of generating quality responses, but these are not consistently produced. Commercial Platform 3 performed more poorly overall, with particularly low median scores in clinical relevance and organization, underscoring the risks of integrating tools that lack veterinary-specific optimization. These findings collectively highlight that the use of an LLM does not guarantee accuracy or clinical utility; platform-specific training data, prompting methods, and error-checking strategies play a decisive role.

This study has several limitations. The evaluation was limited to veterinary oncology records, which may not fully represent the broader spectrum of veterinary specialties. While using an LLM as an automated judge offers advantages in terms of scalability and consistency, several limitations warrant consideration. First, the LLM judge may exhibit inherent biases present in its training data, potentially favoring certain writing styles or organizational patterns over others, regardless of clinical accuracy.[20] Second, despite the structured prompting and criterion-based approach, the evaluation remains fundamentally based on pattern matching rather than true clinical comprehension, which could lead to overlooking subtle but clinically significant errors or nuances that a domain expert would identify.

Additionally, the LLM's assessment of "clinical relevance" relies on statistical patterns rather than genuine medical judgment, potentially misweighting the importance of certain findings. The model may also demonstrate inconsistent performance across different medical subspecialties or rare conditions that were underrepresented in its training data. To mitigate these limitations, we employed structured prompting with explicit evaluation criteria, included the full source text in each evaluation to ground assessments in actual content rather than the model's parametric knowledge, enforced JSON schema validation to ensure consistent output formatting, and used weighted scoring developed in consultation with veterinary clinicians to align algorithmic assessment with clinical priorities.

Future work should validate these automated assessments against expert veterinary clinician evaluations to establish inter-rater reliability and identify systematic discrepancies between algorithmic and human judgment. Despite these limitations, the LLM judge provides a reproducible, scalable method for initial comparative assessment across platforms, particularly useful for identifying relative performance differences when absolute accuracy assessment would require extensive expert review, picking up mistakes that even trained clinicians could miss.

The dataset of 42 records provided a sufficient base for this assessment, but it represents a focused and relatively small sample. Future work should incorporate larger, more diverse datasets, expand the range of clinical prompts, and adopt blinded evaluation protocols.

The findings of this study have practical implications for veterinary medicine. They emphasize the necessity of rigorous validation before incorporating LLM-based tools into clinical workflows. While LLMs offer significant potential to support efficiency and decision-making, practitioners must remain cautious about their limitations, particularly with respect to factual accuracy and reliability. The observed variability across platforms underscores that critical evaluation of outputs is essential, and that tool selection should be guided by clinical context and performance data. Looking ahead, standardized evaluation metrics, tools and datasets will be crucial for integrating LLMs safely and effectively into veterinary practice.

**Conclusion:**
This comparative evaluation highlights that not all platforms perform equally in veterinary medicine. While commercially available tools demonstrated variability and inconsistency, there was clear benefit in a veterinary-trained model pipeline that produced outputs that were more accurate, organized, and clinically relevant. These findings emphasize that domain-specific data and training pipelines are critical for ensuring reliability and clinical utility.

As LLMs continue to be integrated into veterinary practice, careful validation, transparency in model development, and critical appraisal of outputs remain essential. Future work should expand beyond oncology, incorporate a broader range of clinical prompts, and explore standardized evaluation frameworks to guide responsible adoption of AI in veterinary healthcare.

**Figures:**
**Figure 1.** Boxplot comparison of average weighted scores across three LLM-powered platforms. The veterinary-trained tool (Product 1) achieved the highest performance, with consistently high median scores and a narrow interquartile range, reflecting both accuracy and stability. In contrast, Commercial Products 2 and 3 demonstrated markedly lower median scores and greater variability, indicating less reliable performance across evaluation categories.

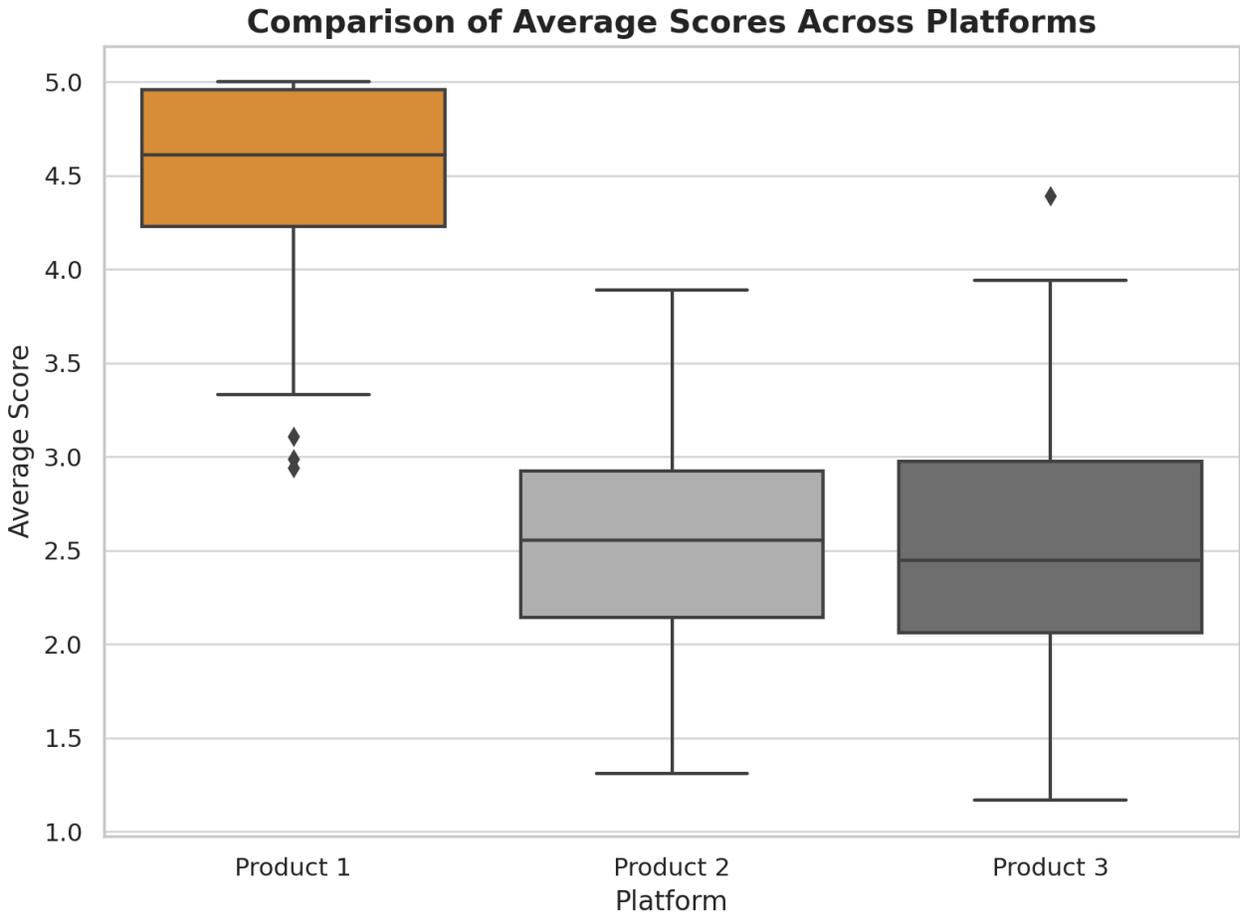

**Figure 2.** Average scores across five evaluation categories: Chronological Order, Clinical Relevance, Completeness, Factual Accuracy, and Organization. Product 1 consistently achieved the highest scores across all categories, with median values of 4.0–5.0 and average scores above 4.0. In contrast, Commercial Products 2 and 3 demonstrated substantially lower performance.

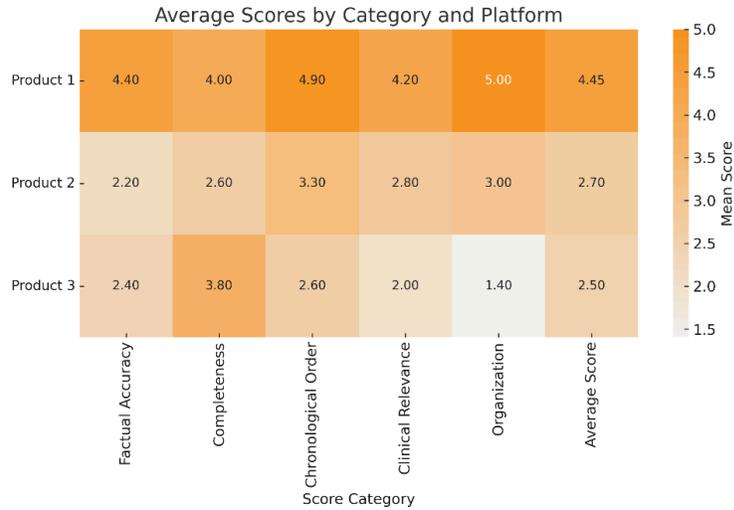

**Figure 3.** Boxplot distribution of scores by category and platform, illustrating variability in output quality across evaluation criteria. Product 1 showed high median scores with narrow interquartile ranges, indicating both superior and more consistent performance. Commercial Product 2 exhibited wider distributions with moderate medians, while Commercial Product 3 displayed the lowest medians and broader variability, particularly in Organization and Clinical Relevance.

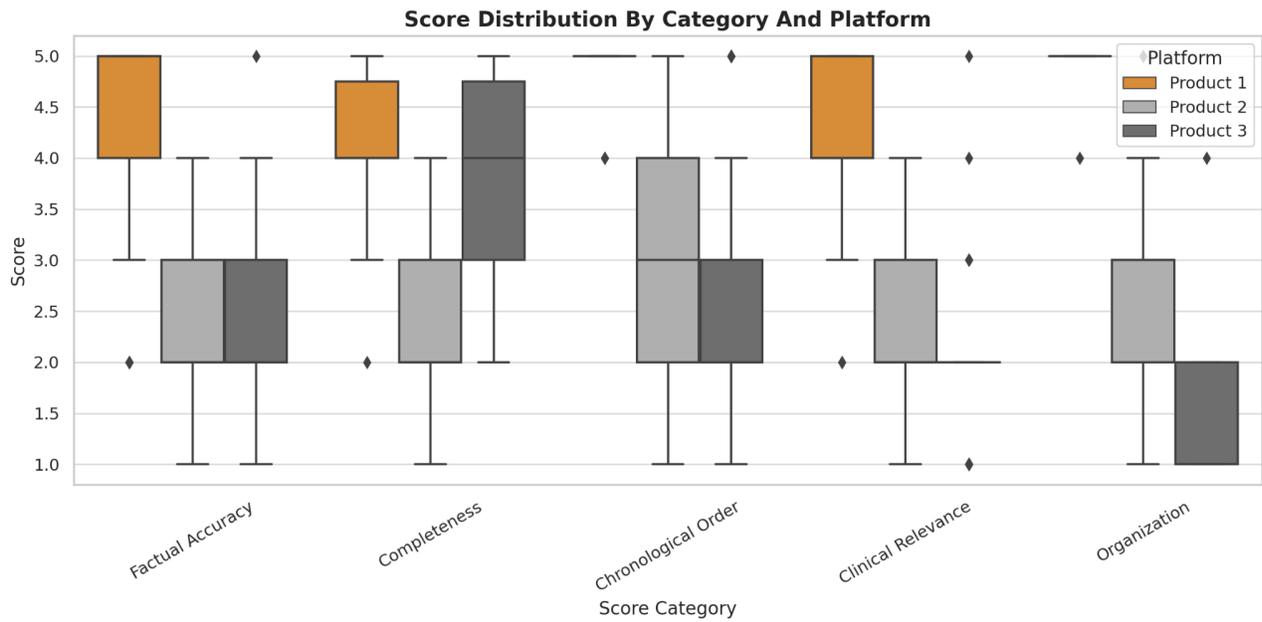

**Figure 4.** Standard deviation of rubric-based scores assigned by the LLM evaluator across three independent grading runs for each summarization platform. Lower standard deviation indicates greater consistency of the grading tool in assessing the same outputs. Product 1 (Hachiko) exhibited the highest reproducibility with a standard deviation of 0.015 for Average Score, compared to 0.088 for Product 2 and 0.034 for Product 3. Similar trends were observed across individual categories, including Factual Accuracy (0.036 vs. 0.113 and 0.143) and Chronological Order (0.036 vs. 0.172 and 0.084), confirming the internal reliability of the LLM-as-a-judge approach for evaluating clinical summaries in veterinary medicine.

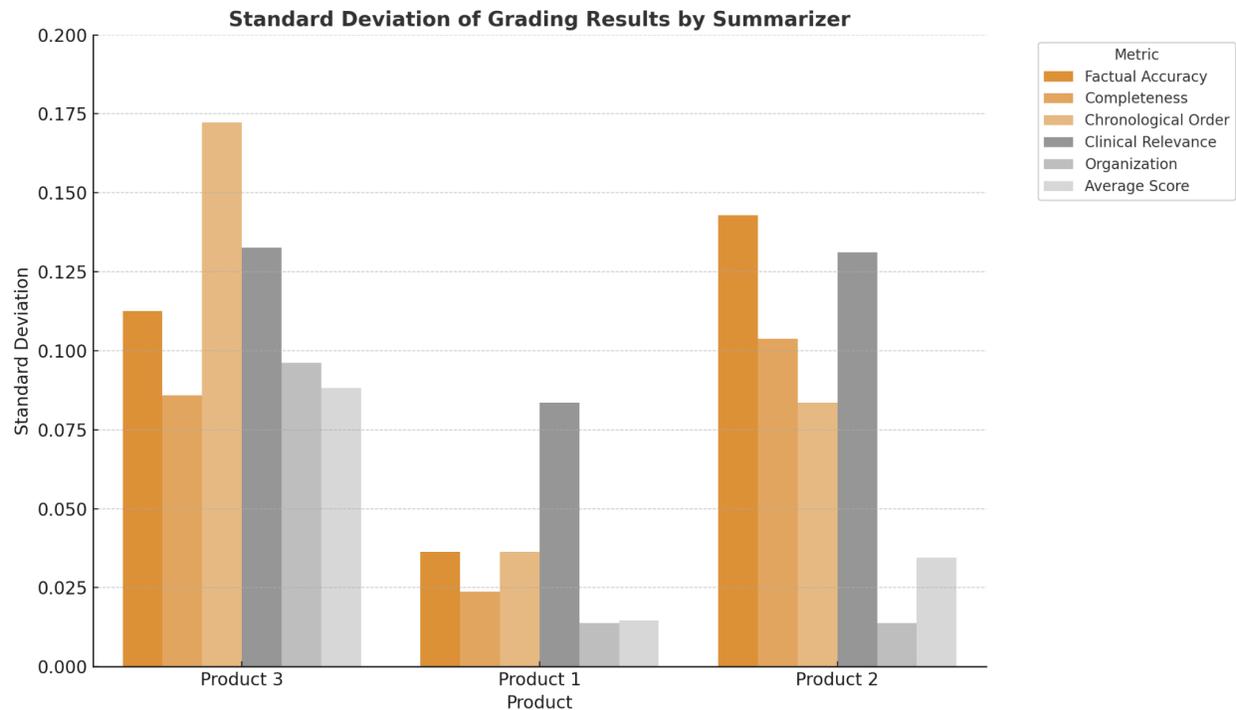